\documentclass[10pt,twocolumn,letterpaper]{article}

\usepackage{cvpr}
\usepackage{times}
\usepackage{epsfig}
\usepackage{graphicx}
\usepackage{amsmath}
\usepackage{amssymb}

\usepackage[breaklinks=true,bookmarks=false]{hyperref}

\cvprfinalcopy % *** Uncomment this line for the final submission

\def\cvprPaperID{****} % *** Enter the CVPR Paper ID here

\begin{document}

%%%%%%%%% TITLE
\title{Clustering and Classification Networks}

\author{Jin-mo Choi\\ 
UST\\
{\tt\small choijm80@etri.re.kr}
\and
Sang Jun Park\\
ETRI\\
{\tt\small sangjoon@etri.re.kr}
}

\maketitle
%\thispagestyle{empty}

%%%%%%%%% ABSTRACT
\begin{abstract}

In this paper, we will describe a network architecture that demonstrates high performance on various sizes of datasets. To do this, we will perform an architecture search by dividing the fully connected layer into three levels in the existing network architecture. The first step is to learn existing CNN layer and existing fully connected layer for 1 epoch. The second step is clustering similar classes by applying L1 distance to the result of Softmax. The third step is to reclassify using clustering class masks. We accomplished the result of state-of-the-art by performing the above three steps sequentially or recursively. The technology recorded an error of 11.56\% on Cifar-100

% and an error of xx.xx\% on Image net. 
\end{abstract}

%%%%%%%%% BODY TEXT
\section{Introduction}

Since AlexNet\cite{cite00} won ILSVRC 2012, classification in the field of computer vision has developed rapidly. Alex Net used ReLU to deeper neural networks and Dropout to mitigate overfitting problems. CNN and pooling layer were used for efficient feature representation and fully connected layer was used for label mapping. AlexNet has made the neural network the most noticeable technology in machine learning.

Another innovation of Classification is from ResNet\cite{cite01, cite02}. ResNet has made it possible to configure depths of more than 1000 layers using an identity skip connection. Since then, ResNet has inspired many researchers to efficiently extend the technology. Wide residual networks have relieved the vanishing gradient problem by increasing channels instead of depth, and ResNeXt\cite{cite04} used cardinality to express feature diversity. PolyNet\cite{cite05} uses the recursive property of skip connection to increase the representaion power.

ResNet has also been a good candidate for understanding the characteristics of the network architecture. \cite{cite28} has shown the impact of identity skip connections as an unraveled view, showing that the residual network behaves like an ensemble of shallow networks instead of a single deep network. Highway and residual networks learn unrolled iteration networks show that successive residual blocks within a stage refine existing features. This means that the feature is learned in stages, and it is based on the interpretation of the signal processing of the Deep convolutional framelet\cite{cite08}, which improves the performance.

NasNet\cite{cite30} conducted an architecture search using Reinforcement learning. The technology has set up various types of convolution layer, pooling layer, and identity skip connection as inputs and automatically designed the network architecture using RNN controller. In order to search the network architecture quickly, it is learned by the unit of Normal Cell and Reduction Cell. The technology achieves excellent accuracy compared to the network architecture designed by human.

DenseNet\cite{cite35} used a concatenate technique called dense connectivity. This method increased feature variation of subsequent layers by reusing features of other layers and facilitated learning by directly propagating the gradient of each layer from the loss function. DenseNet on the ImageNet dataset is able to learn efficiently with fewer parameters compared to ResNet. This method also achieved impressive performance improvements using fewer parameters on smaller datasets such as Cifar-10, Cifar-100, and SVHN. The technology has been applied to network architecture for real-time mobile devices such as Pelee\cite{cite36}.

Although various technology approaches have been made in this way, the technological progress of network architecture is gradually slowing down. Although research on reducing the number of parameters while preserving the performance of the existing network architecture, such as prunning, has been continuously carried out, the number of publications that increase the performance of the network architecture itself has decreased dramatically. This is because the conventional methods do not raise the accuracy anymore. We assume that one of the fundamental reasons for this phenomenon is that we do not consider a network architecture that is specific to datasets. If we design a network architecture that is specific to a particular dataset, we can improve accuracy. The problem is that designing a specialized network architecture in the usual way is very time-consuming to invest, whether using automated machine learning or human designing.

To solve these problems, this paper proposes a technique to apply automated machine learning using only fully connected layer using existing network architecture. Since most recently developed technologies use Average pooling as the final result of CNN layer, applying automated machine learning to fully connected layer is very efficient in memory consumption and learning time. This is because unlike VGG, most network architectures are not bottlenecks in a fully connected layer. Also, it can be applied easily by using the existing CNN layer technology which is responsible for feature representation. Moreover, it is expected to help design a modular network architecture.

Another inspiration of the proposed technique is from the human neuron structure. The human cerebral cortex responds locally to certain objects and has an anatomically parallel connection structure. What this means is that the neurons that respond to a particular object are part of region. When a fully connected layer, such as an existing neural network, is not specific to a particular object, it is configured to map directly to the target or to affect all targets when the output of the previous fully connected layer is connected to the input of the latter fully connected layer. This forces the computation of highly complex decision boundaries to the fully connected layer.

To solve this, we clustering targets with similar characteristics. This can dramatically reduce the number of targets. Then, the target included in each clustering is classified. This turns the problem of computing a single complex decision boundary into a problem of calculating two simple decision boundaries. To do this, we share the role in 3 steps as follows.

The first step is to classify the dataset in the usual way using the existing network architecture. Since we extend the fully connected layer to improve the performance of the network architecture, the output of the CNN layer used in the process is limited to the model with average pooling. Since the result of the process is used to measure the similarity between labels in the following process, 1 epoch is performed considering all input data.

The second step is to measure the similarity between the labels using the results of the first step, and then use them to generate new clustering labels and learn them. The output of the fully connected layer used in the process is switched to the output of the clustering label and fine-tuned with the CNN layer.

The third step creates a new fully connected layer that separates the labels in each cluster using the clustering information generated in the second step. We call it Clustered Classification, which consists of 2 fully connected layers and a ReLU layer, and finally a mask used to clustering labels in step 2.

The contribution of this paper is as follows.

First, I will explain why we need a specialized hierarchy for a particular object that is claimed by neuro science. This will approach the complex decision boundary problem as a simplified problem.

Second, we identify the structure of the dataset and propose a clustering and classification structure specific to the dataset. We analyze the structure of the dataset in the first step to efficiently analyze and learn the dataset, learn clustering labels in the second step and classify the labels in the cluster in the third step.

Third, it will be proved that it can be easily applied to various existing network architectures to improve performance.

Fourth, we will demonstrate the performance of state of the art on cifar-10, cifar-100, and ImageNet through experiments.

\section{Clustering and Classification Network}

According to Neuro science, humans respond to the area of the cerebral cortex for a specific object. In addition, when a human neuron connection is scanned, it can be seen that the brain is connected in parallel from the center of the brain to the cortex. This allows us to infer that the human neurons are specialized learners for the target.

We hypothesized that these neuron structures hierarchically divide the target, such as the hierarchical temporal memory \cite{cite37} (HTM). When each level is defined as a level, the advantage of this structure is that the calculation complexity can be dramatically reduced by dividing the decision boundary by level for a very large number of objects.

\begin{figure}[t]
\begin{center}
   \includegraphics[width=0.8\linewidth]{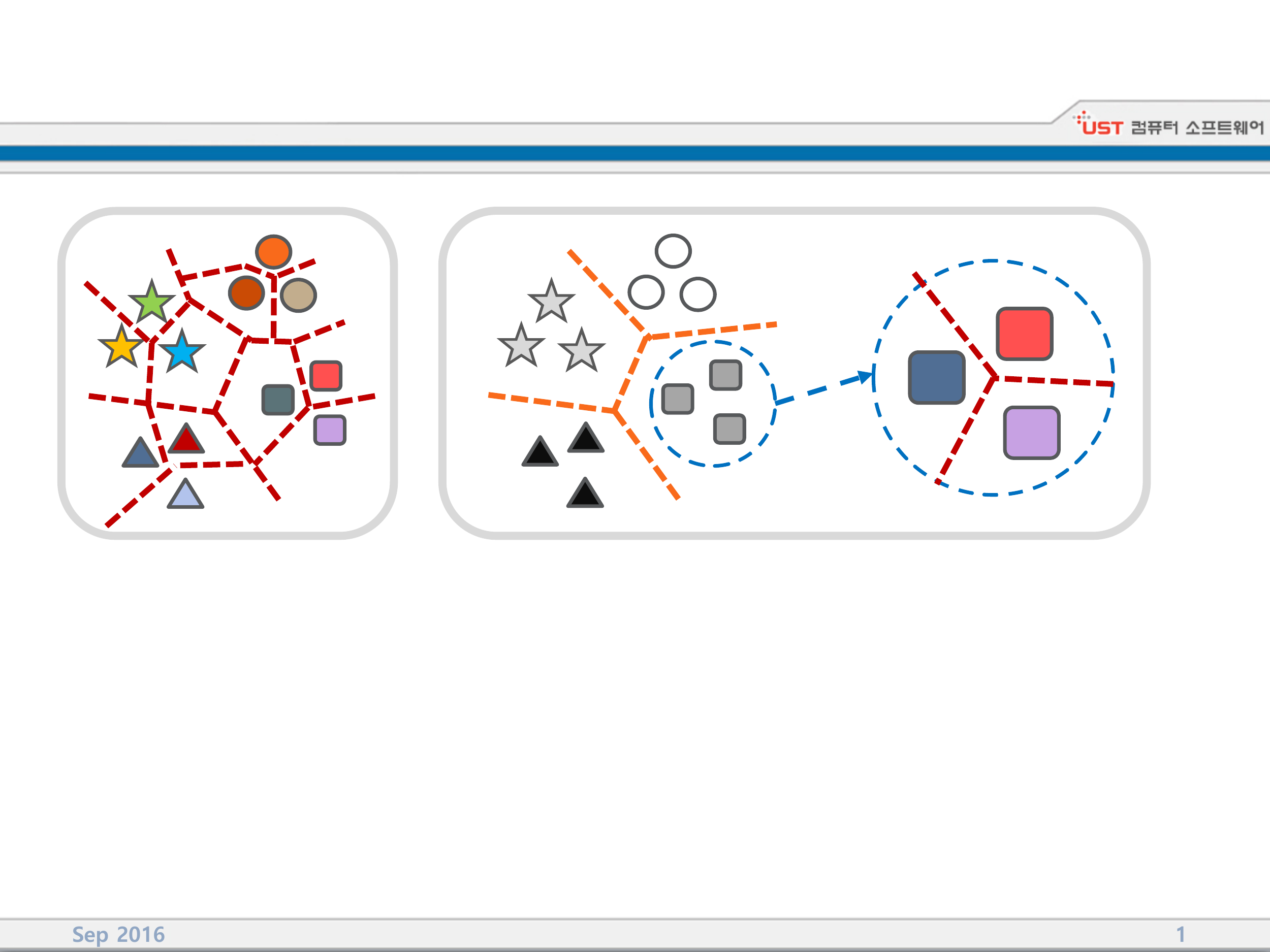}
\end{center}
   \caption{one-level target mapping and two-level target mapping}
\label{fig:long}
\label{fig:onecol}
\end{figure}

Figure 1 (a) shows one level target mapping that we commonly use in neural networks. Figure 2 (b) shows the hierarchical target mapping proposed by the neural network. Figure 1 (a) requires a very complex decision boundary, whereas Figure 1 (b) requires a simple decision boundary. This means that when classifying a very large number of objects using 1 level target mapping, it requires excessive computational complexity for a fully connected layer.

\begin{figure}[t]
\begin{center}
   \includegraphics[width=0.8\linewidth]{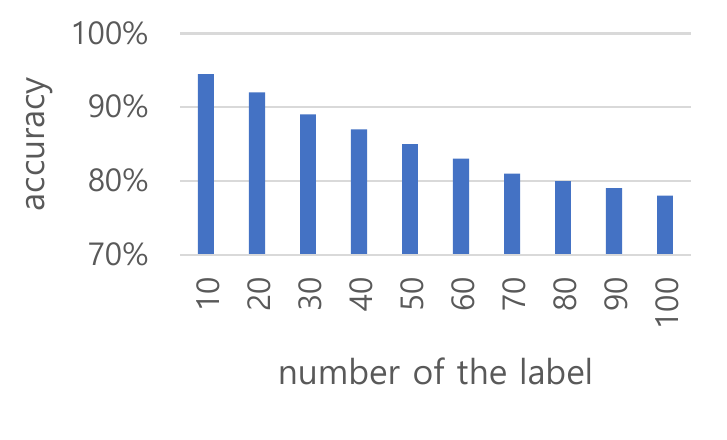}
\end{center}
   \caption{Comparison of accuracy when label number is limited using Cifar-100}
\label{fig:long}
\label{fig:onecol}
\end{figure}

Figure 2 compares the accuracy of Cifar-100 when the number of labels is limited. The network architecture used ResNet-50 and only the number of output of the fully connected layer was modified according to the number of labels. Label was randomly selected on cifar-100. From Figure 2, we can see that the number of labels has a serious impact on the performance of the classification.

\subsection{Clustering}
We were affected by the Figure 2 and needed a new target mapping layer. To do this, we needed a new rule for clustering at each level.

\begin{figure}[t]
\begin{center}
   \includegraphics[width=0.8\linewidth]{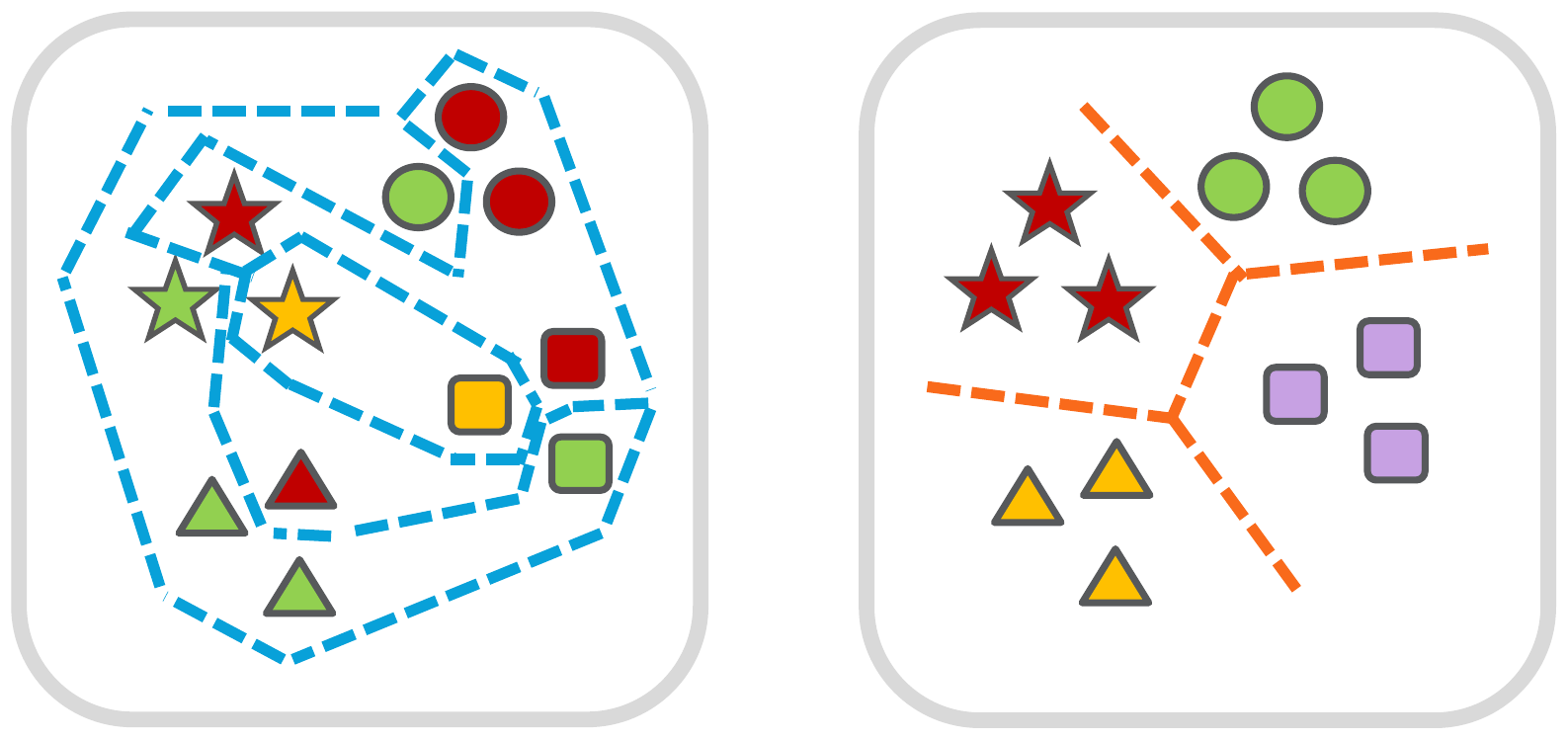}
\end{center}
   \caption{Decision boundary according to clustering}
\label{fig:long}
\label{fig:onecol}
\end{figure}

Figure 3 shows the difference of decision boundary due to incorrect clustering and correct clustering. This means that the result of erroneous clustering has a serious adverse effect on performance. To solve this problem, we applied the following simple and superior clustering technique.

\begin{equation}\label{eq:exp1}
\hat{Y}_{L}=\sum_{l}^{l_{num}} \hat{y}_{il}
\end{equation}

\begin{equation}\label{eq:exp2}
\hat{Y}_{L}= \{\hat{y}_{il}|\hat{y}_{il}\in \mathbf {\hat{Y}_{L}}, 1 \leq i\leq i_{num}\} 
\end{equation}

\begin{equation}\label{eq:exp3}
S = argmax_{l}(\hat{Y}_{L}), \hat{y}_{l} > trsd 
\end{equation}

$\hat{y}_{il}$ is the value predicted by the neural network. $l$ denotes the clustered label of each level, and since the label is not clustering at the beginning, the value of clustered label and label are the same. $\hat{y}_{il}$ is the predicted value for the $i$-th label defined by the true label $l$. $\sum_{i}^{i_{num}} \hat{y}_{il}=1$ because it is the result of softmax. $\hat{Y}_{L}$ indicates the relationship of the predicted value with the l-th label to the similar label. For example, the $\hat{y}_{l}$ value in Equation 2 represents the relationship of the $i$-th label to true $l$-label. We used clustering when the corresponding value was high as shown in Equation 3.

\begin{figure}[t]
\begin{center}
   \includegraphics[width=0.8\linewidth]{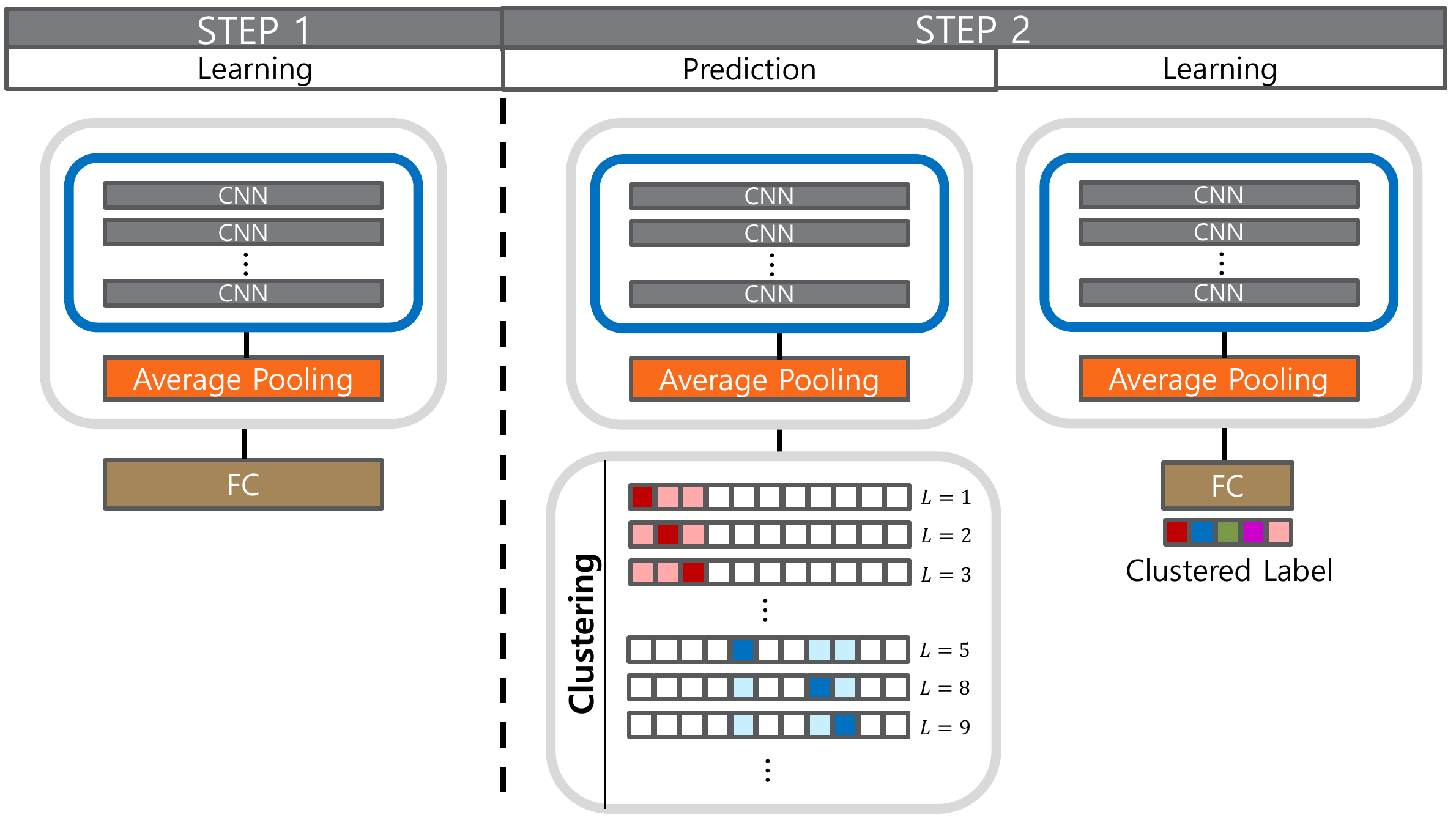}
\end{center}
   \caption{Step1 and step2 algorithm flow chart}
\label{fig:long}
\label{fig:onecol}
\end{figure}

\begin{figure}[t]
\begin{center}
   \includegraphics[width=0.8\linewidth]{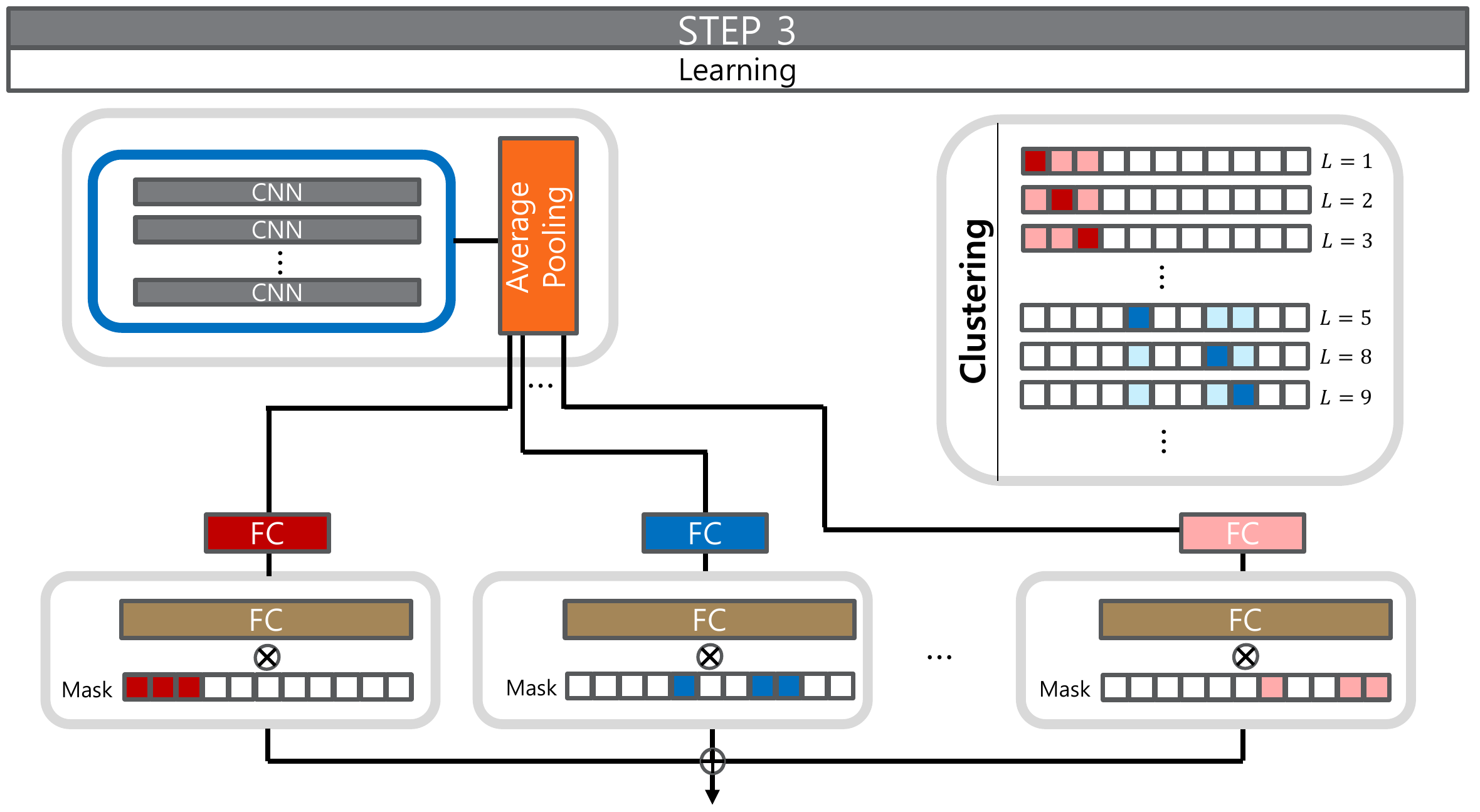}
\end{center}
   \caption{Step3 algorithm flow chart}
\label{fig:long}
\label{fig:onecol}
\end{figure}

Figure 4 shows the process of creating Clustering and Clustered Label in Step2 using Network architecture of Step1. Clustering is performed in the prediction process using the weight learned in Step 1. For example, L = 1, L = 2 and L = 3 are assigned Clustered Label L '= 1 and L = 5, L = 8 and L = 9 are assigned Clustered Label L' = 2. This redefined Label (Clustered Label) is subjected to a fine-tuning process in Step 2. At this time, the weight of the Fully connected layer used in step 1 is discarded. We performed only one epoch in Step 1 to calculate the relationship of labels that are not distorted by the network architecture. In Step 2, we performed epoch until convergence. Clustered labels contribute to dramatically reducing the number of labels and lowering the complexity of decision boundaries.

\subsection{Classification}
We connect the fully connected layer in parallel to classify the original label of the Clustered Label. For this reason, when there are many input channels of Fully connected layer like VGG, the memory cost becomes high. However, thanks to Average pooing, the parallel connection cost of the Fully connected layer is relatively low compared to the CNN layer. As shown in Figure 4, step 3 consists of a two-layer fully connected layer with each fully connected branch. ReLU is inserted between the fully connected layers. The output of the first fully connected layer increases memory efficiency by allocating fewer channels than the second fully connected layer. The output of the second fully connected layer is equal to the number of original labels and is used as a mask by recycling the clustering information of step 2. Step 3 fine-tunes the learned weight of Step 2 and discards the fully connected layer used in Step 2. The process is performed until convergence. This allows the brunch of each fully connected layer to compute decision boundaries specific to clustered labels and to perform the classification of each branch.

\subsection{Multi-level target mapping}
Our method varies the depth of the fully connected layer depending on the number of steps2 and step3. If you want to create a decision boundary in two steps like the two level target mapping in Figure 5, you only need to perform Step 2 and Step 3 once. On the other hand, if the clustered labels are recursively clustering by performing Step 2 and Step 3 N times recursively, N + 1 target level mapping is possible. In the case of two level target mapping, the number of clustering at each level is relatively large, while the error propagation is stable. On the other hand, in the case of multi-level target mapping, the number of clustering at each level is relatively small, which makes it possible to construct a decision boundary with a low computational complexity, but there is a risk that error propagation is less and the proportion of memory usage increases. We will compare the performance of each configuration through experiments.

\begin{figure}[t]
\begin{center}
   \includegraphics[width=0.8\linewidth]{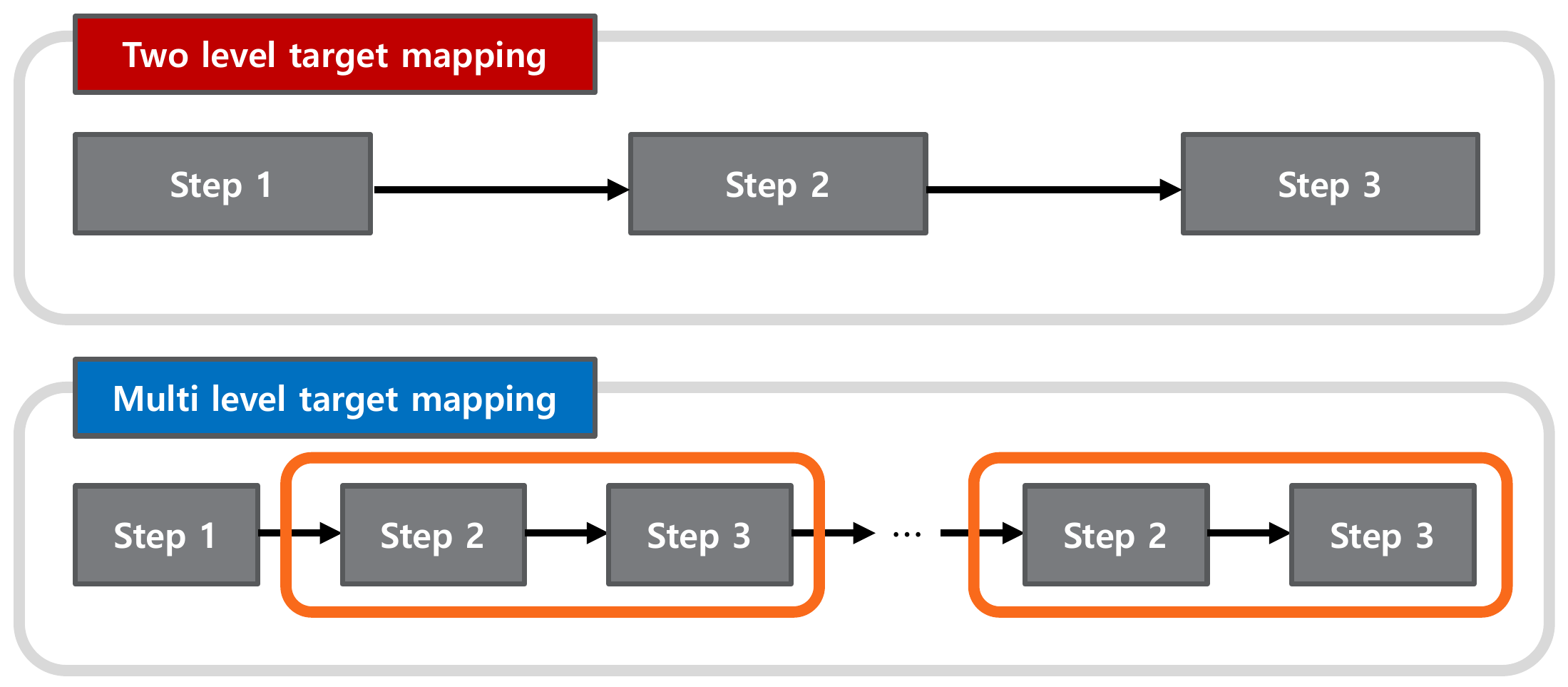}
\end{center}
   \caption{Two-level target mapping and Multi-level target mapping}
\label{fig:long}
\label{fig:onecol}
\end{figure}

\section{Experiments}
\subsection{Implements}
% In order to measure the versatility of CNCNet, we experimented with various existing network architectures. Initialization of the weight of the network architecture and various hyper-parameters, optimization, etc., used the existing values of the existing models. However, a model that requires many epochs, such as shakeshake regularization, has also experimented with short epochs. In our proposed network architecture, the model name is prefixed with CNC. The input channel of CNCNet's first fully connected layer is set four times larger than the input channel of the second fully connected layer. Framwork uses pytorch and the test environment is Intel® CPU E5-2650 v4 @ 2.20GHZ and NVidia Tesla P40 2GPU. The experimental datasets are Cifar-10, Cifar-100, and ImageNet.

In order to measure the performance of CnCNet, we experimented with applying the existing network architecture and CnCNet to the network. The model used is ShakeShake regularization and is simply referred to as SSR in the table. The model to which CnC is applied is denoted as SSR-CnC. Also, for quick experimental results, we applied only 300epochs to the comparative model. The dataset is cifar-100.
The input channel of CNCNet's first fully connected layer is set four times larger than the input channel of the second fully connected layer. Framwork uses pytorch and the test environment is Intel® CPU E5-2650 v4 @ 2.20GHZ and NVidia Tesla P40 2GPU.

% \subsection{Cifar-10}
% The Cifar-10 consists of 32 × 32 color images for classifying 10 classes. It is divided into 50000 training images and 10000 test images. The dataset is a standard data augmentation using horizontal flip and random crop.

% \begin{table}
% \begin{center}
% \begin{tabular}{|l|c|c|c|c|}
% \hline
% Network & Depth-k & \# Params & Top-1 error\\
% \hline\hline
% ResNet & 110 & 1.7M & 6.43 \\
% ResNet & 1202 & 10.2M & 7.93 \\
% \hline
% ResNet-CNC & 110 & & \\
% ResNet-CNC & 1202 & & \\
% \hline\hline
% pre-ResNet & 110 & 1.7M & 6.37 \\
% pre-ResNet & 1001 & 10.2M & 4.92 \\
% \hline
% pre-ResNet-CNC & 110 & &  \\
% pre-ResNet-CNC & 1001 & &  \\
% \hline\hline
% WRN & 16-8 & 11.0M & 4.27 \\
% WRN & 28-10 & 36.5M & 4.00 \\
% \hline
% WRN-CNC & 16-8 & & \\
% WRN-CNC & 28-10 & & \\
% \hline\hline
% SSR@1800 & 26-32 & & 3.67 \\
% SSR@1800 & 26-64 & & 2.86 \\
% \hline
% SSR-CNC@1800 & 26-32 & & \\
% SSR-CNC@1800 & 26-64 & & \\
% \hline\hline
% SDR@1800 & 272 & 26.0M & 2.67 \\
% \hline
% SDR-CNC@1800 & 272 &  &  \\
% \hline
% \end{tabular}
% \end{center}
% \caption{Cifar-10 Results.}
% \end{table}

\subsection{Cifar-100}
The Cifar-100 consists of 32 × 32 color images for classifying 100 classes. It is divided into 50000 training images and 10000 test images. The data dataset was applied to standard data augmentation like cifar-10

Table 1 shows very low top-1 error even though the parameter is only 3.1M. This is the result of state-of-the-art in a network architecture built on Cifar-100. Existing state-of-the-art shake-shake regularization and shake-drop regularization achieved 15.85\% and 13.99\% on parameters of 34.4M and 26.0M, respectively. Our proposed network architecture achieved a result of 11.56\% over 3.1M.

\subsection{Conclusion}
In this paper, we have proposed a new way to efficiently perform a network architecture search of a fully connected layer, inspired by human biological neurons. The technology has created a network architecture optimized for datasets with relatively low time consumption. Experimental results show that cifar-100 performs very well, and we expect it to perform well on large datasets such as ImageNet.

\begin{table}
\begin{center}
\begin{tabular}{|l|c|c|c|c|}
\hline
Network & Depth-k & \# Params & Top-1 error\\
% \hline\hline
% ResNet & 110 & 1.7M & 25.16 \\
% ResNet & 1202 & 10.2M & 27.82 \\
% \hline
% ResNet-CNC & 110 & & \\
% ResNet-CNC & 1202 & & \\
% \hline\hline
% pre-ResNet & 164 & 1.7M & 24.33 \\
% pre-ResNet & 1001 & 10.2M & 22.71 \\
% \hline
% pre-ResNet-CNC & 110 & &  \\
% pre-ResNet-CNC & 1001 & &  \\
% \hline\hline
% WRN & 16-8 & 11.0M & 20.43 \\
% WRN & 28-10 & 36.5M & 19.25 \\
% \hline
% WRN-CNC & 16-8 & & \\
% WRN-CNC & 28-10 & & \\
\hline\hline
Res-SSR@300 & 26-32 & 2.9M & 23.51 \\
% SSR@1800 & 29 & 34.4M & 15.85 \\
\hline
Res-SSR-CNC@300 & 26-32 & 3.1M & \textbf{11.56}\\
% SSR-CNC@1800 & 29 & &  \\
% \hline\hline
% SDR@300 & 272 & 26.0M & 14.90 \\
% SDR@1800 & 272 & 26.0M & 13.99 \\
% \hline
% SDR-CNC@300 & 272 &  &  \\
% SDR-CNC@1800 & 272 &  &  \\
\hline
\end{tabular}
\end{center}
\caption{Cifar-100 Results.}
\end{table}

% \subsection{ImageNet}
% ImageNet is a dataset for classifying 1000 classes, consisting of more than one million images and training images and 50,000 test images. We experimented with 224x224 as input image.

% % \begin{figure*}
% % \begin{center}
% % \fbox{\rule{0pt}{2in} \rule{.9\linewidth}{0pt}}
% % \end{center}
% %   \caption{Example of a short caption, which should be centered.}
% % \label{fig:short}
% % \end{figure*}

% \begin{table}
% \begin{center}
% \begin{tabular}{|l|c|c|c|c|}
% \hline
% Network & Depth-k & \# Params & Top-1 error\\
% \hline\hline
% ResNet & 152 &  & 23.0 \\
% ResNet & 200 &  & 21.7 \\
% \hline
% ResNet-CNC & 152 & & \\
% ResNet-CNC & 200 & & \\
% \hline\hline
% ResNeXt & 101 & & 20.4 \\
% \hline
% ResNeXt-CNC & 101 & &  \\
% \hline\hline
% SENet &  &  & 18.68 \\
% \hline
% SENet-CNC &  & & \\
% \hline
% \end{tabular}
% \end{center}
% \caption{ImageNet Results.}
% \end{table}

%-------------------------------------------------------------------------

% List and number all bibliographical references in 9-point Times,
% single-spaced, at the end of your paper. When referenced in the text,
% enclose the citation number in square brackets, for
% example~\cite{Authors14}.  Where appropriate, include the name(s) of
% editors of referenced books.

{\small
\bibliographystyle{ieee}
\bibliography{egbib}
}

\end{document}